\pdfoutput=1
\documentclass[10pt,twocolumn,letterpaper]{article}
\usepackage{cvpr}
\usepackage{times}
\usepackage{epsfig}
\usepackage{graphicx}
\usepackage{xcolor}
\usepackage{epstopdf}
\usepackage{amsmath}
\usepackage{amssymb}
\usepackage{tabularx}
\usepackage{bm}
\usepackage{caption}
\usepackage{subcaption}
\usepackage{placeins}
\usepackage{grffile}

\usepackage[pagebackref=true,breaklinks=true,letterpaper=true,colorlinks,bookmarks=false]{hyperref}
\usepackage{cleveref}

\cvprfinalcopy 
\def\cvprPaperID{3198} 

\ifcvprfinal\fi

\newcommand{\bx}{\mathbf{x}}

\newcommand{\z}{{\ensuremath{{z}}}\xspace}
\newcommand{\y}{{\ensuremath{{y}}}\xspace}
\newcommand{\x}{{\ensuremath{{x}}}\xspace}

\newcommand{\argmin}{\operatornamewithlimits{argmin}}

\newcommand{\E}{\operatornamewithlimits{\mathbb{E}}}

\newcommand{\deflen}[2]{%
    \expandafter\newlength\csname #1\endcsname
    \expandafter\setlength\csname #1\endcsname{#2}%
}

\newcommand\blfootnote[1]{%
  \begingroup
  \renewcommand\thefootnote{}\footnote{#1}%
  \addtocounter{footnote}{-1}%
  \endgroup
}

\title{%
Improved Texture Networks: Maximizing Quality and Diversity in\\
 Feed-forward Stylization and Texture Synthesis}
\author{Dmitry Ulyanov\\
Skolkovo Institute of Science and Technology \& Yandex\\
{\tt\small dmitry.ulyanov@skoltech.ru}
\and
Andrea Vedaldi\\
University of Oxford\\
{\tt\small vedaldi@robots.ox.ac.uk}
\and
Victor Lempitsky\\
Skolkovo Institute of Science and Technology\\
{\tt\small lempitsky@skoltech.ru}
}
\begin{document}
\maketitle
\begin{abstract}
   The recent work of Gatys~\etal, who characterized the style of an image by the statistics of convolutional neural network filters, ignited a renewed interest in the texture generation and image stylization problems. While their image generation technique uses a slow optimization process, recently several authors have proposed to learn generator neural networks that can produce similar outputs in one quick forward pass. While generator networks are promising, they are still inferior in visual quality and diversity compared to generation-by-optimization. In this work, we advance them in two significant ways. First, we introduce an instance normalization module to replace batch normalization with significant improvements to the quality of image stylization. Second, we improve diversity by introducing a new learning formulation that encourages generators to sample unbiasedly from the Julesz texture ensemble, which is the equivalence class of all images characterized by certain filter responses. Together, these two improvements take feed forward texture synthesis and image stylization much closer to the quality of generation-via-optimization, while retaining the speed advantage. \blfootnote{The source code is available at \url{https://github.com/DmitryUlyanov/texture_nets}}
\end{abstract}

\section{Introduction}

The recent work of Gatys~\etal~\cite{Gatys15,Gatys16}, which used deep neural networks for texture synthesis and image stylization to a great effect, has created a surge of interest in this area. Following an earlier work by Portilla and Simoncelli~\cite{Portilla00}, they generate an image by matching the second order moments of the response of certain filters applied to a reference texture image. The innovation of Gatys~\etal is to use non-linear convolutional neural network filters for this purpose. Despite the excellent results, however, the matching process is based on local optimization, and generally requires a considerable amount of time (tens of seconds to minutes) in order to generate a single textures or stylized image.

\begin{figure}[t]
    \centering
    \hspace*{-0.3mm}\begin{subfigure}[b]{0.32\linewidth}
        \includegraphics[width=\linewidth]{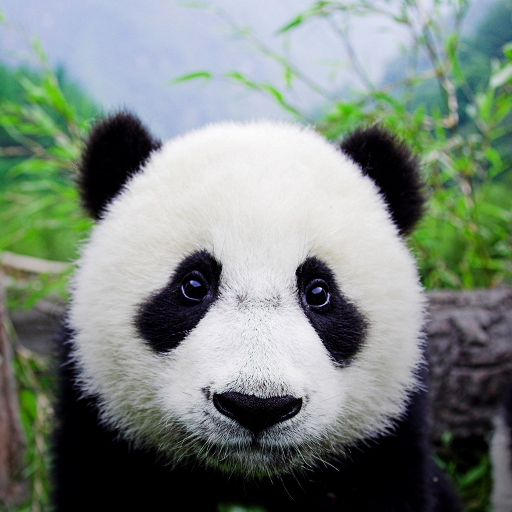}
        \vspace*{-6mm}\caption{Panda.}
    \end{subfigure}
    \hspace*{2mm}\begin{subfigure}[b]{0.32\linewidth}
        \includegraphics[width=\linewidth]{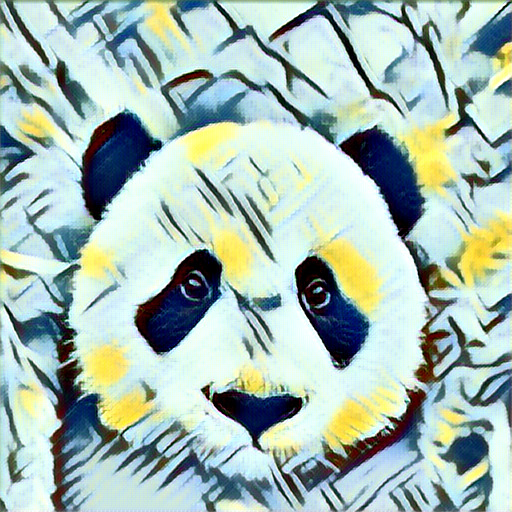}
        \vspace*{-6mm}\caption{}
    \end{subfigure}
    \begin{subfigure}[b]{0.32\linewidth}
        \includegraphics[width=\linewidth]{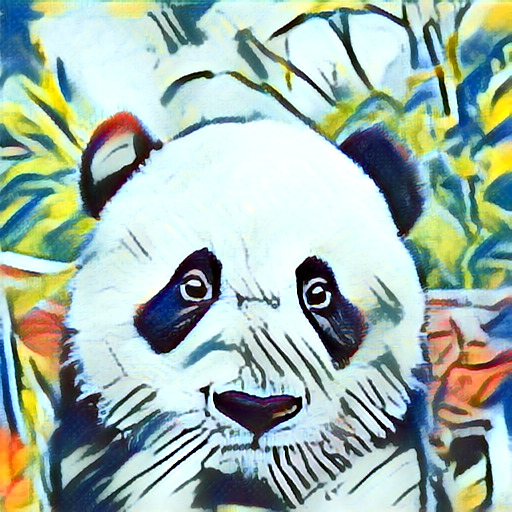}
        \vspace*{-6mm}\caption{}
    \end{subfigure}
    \\
    \hspace*{-0.3mm}\begin{subfigure}[b]{0.32\linewidth}
        \includegraphics[width=\linewidth]{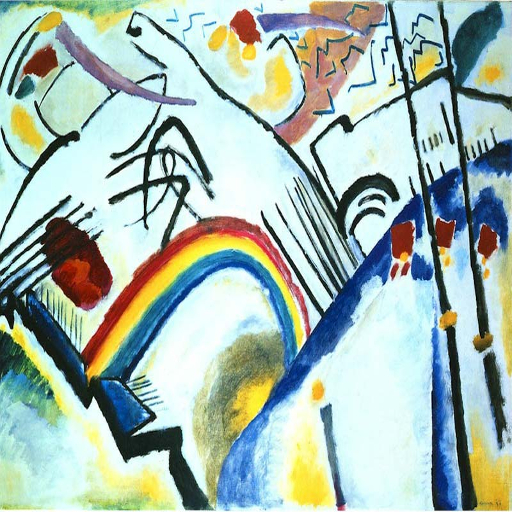}
        \vspace*{-6mm}\caption{Style.}
    \end{subfigure}
    \hspace*{2mm}\begin{subfigure}[b]{0.32\linewidth}
        \includegraphics[width=\linewidth]{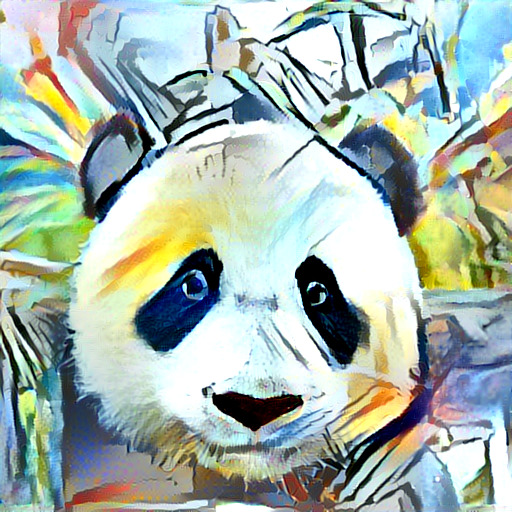}
        \vspace*{-6mm}\caption{}
    \end{subfigure}
    \begin{subfigure}[b]{0.32\linewidth}
        \includegraphics[width=\linewidth]{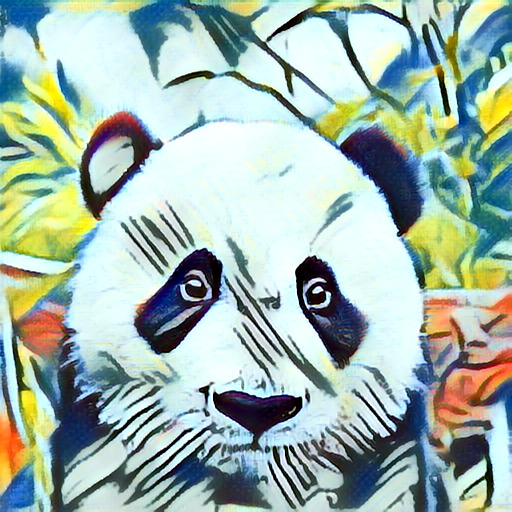}
        \vspace*{-6mm}\caption{}
    \end{subfigure}
    \caption{Which panda stylization seems the best to you? Definitely not the variant (b), which has been produced by a state-of-the-art algorithm among methods that take no longer than a second. The (e) picture took several minutes to generate using an optimization process, but the quality is worth it, isn't it? We would be particularly happy if you chose one from the rightmost two examples, which are computed with our new method that aspires to combine the quality of the optimization-based method and the speed of the fast one. Moreover, our method is able to produce diverse stylizations using a single network.}
\end{figure}

In order to address this shortcoming, Ulyanov~\etal~\cite{Ulyanov16} and Johnson~\etal~\cite{Johnson16} suggested to replace the optimization process with feed-forward generative convolutional networks. In particular, \cite{Ulyanov16} introduced \textit{texture networks} to generate textures of a certain kind, as in \cite{Gatys15}, or to apply a certain texture style to an arbitrary image, as in \cite{Gatys16}. Once trained, such texture networks operate in a feed-forward manner, three orders of magnitude faster than the optimization methods of~\cite{Gatys15,Gatys16}.

The price to pay for such speed is a reduced performance. For texture synthesis, the neural network of \cite{Ulyanov16} generates good-quality samples, but these are not as diverse as the ones obtained from the iterative optimization method of \cite{Gatys15}. For image stylization, the feed-forward results of~\cite{Ulyanov16,Johnson16} are qualitatively and quantitatively worse than iterative optimization. 
In this work, we address both limitations by means of two contributions, both of which extend beyond the applications considered in this paper.

Our first contribution (\cref{s:instance}) is an architectural change that significantly improves the generator networks. The change is the introduction of an \textbf{instance-normalization layer} which, particularly for the stylization problem, greatly improves the performance of the deep network generators. This advance significantly reduces the gap in stylization quality between the feed-forward models and the original iterative optimization method of Gatys~\etal, both quantitatively and qualitatively.

Our second contribution (\cref{s:julesz}) addresses the limited diversity of the samples generated by texture networks. In order to do so, we introduce a new formulation that learns generators that \textbf{uniformly sample the Julesz ensemble}~\cite{zhu00exploring}. The latter is the equivalence class of images that match certain filter statistics. Uniformly sampling this set guarantees diverse results, but traditionally doing so required slow Monte Carlo methods~\cite{zhu00exploring}; Portilla and Simoncelli, and hence Gatys~\etal, cannot sample from this set, but only find individual points in it, and possibly just one point. Our formulation minimizes the Kullback-Leibler divergence between the generated distribution and a quasi-uniform distribution on the Julesz ensemble. The learning objective decomposes into a loss term similar to Gatys~\etal minus the entropy of the generated texture samples, which we estimate in a differentiable manner using a non-parametric estimator~\cite{Kozachenko87}.

We validate our contributions by means of extensive quantitative and qualitative experiments, including comparing the feed-forward results with the gold-standard optimization-based ones (\cref{s:experiments}). We show that, combined, these ideas dramatically improve the quality of feed-forward texture synthesis and image stylization, bringing them to a level comparable to the optimization-based approaches. 
\section{Background and related work}\label{s:background}

\paragraph{Julesz ensemble.} Informally, a \emph{texture} is a family of visual patterns, such as checkerboards or slabs of concrete, that share certain local statistical regularities. The concept was first studied by Julesz~\cite{julesz81textons}, who suggested that the visual system discriminates between different textures based on the average responses of certain image filters.

The work of~\cite{zhu00exploring} formalized Julesz' ideas by introducing the concept of \emph{Julesz ensemble}. There, an image is a real function $x : \Omega \rightarrow \mathbb{R}^3$ defined on a discrete lattice $\Omega = \{1,\dots,H\}\times\{1,\dots,W\}$ and a texture is a distribution $p(x)$ over such images. The local statistics of an image are captured by a bank of (non-linear) filters $F_l : \mathcal{X} \times \Omega \rightarrow \mathbb{R}$, $l=1,\dots,L$, where $F_l(x,u)$ denotes the response of filter $F_l$ at location $u$ on image $x$. The image $x$ is characterized by the spatial average of the filter responses $\mu_l(x) = \sum_{u\in\Omega} F_l(x,u)/|\Omega|$. The image is perceived as a particular texture if these responses match certain characteristic values $\bar \mu_l$. Formally, given the loss function,
\begin{equation}\label{e:loss}
  \mathcal{L}(x) = \sum_{l=1}^L (\mu_l(x) - \bar \mu_l)^2
\end{equation}
the Julesz ensemble is the set of all texture images 
\[
\mathcal{T}_\epsilon = \{ x \in \mathcal{X} : \mathcal{L}(x) \leq \epsilon\}
\]
that approximately satisfy such constraints. Since all textures in the Julesz ensemble are perceptually equivalent, it is natural to require the texture distribution $p(x)$ to be uniform over this set. In practice, it is more convenient to consider the exponential distribution
\begin{equation}\label{e:pjul}
 p(\x) = 
 \frac{
 e^{-\mathcal{L}(\x)/T}
 }{
 \int e^{-\mathcal{L}(\y)/T}\,d\y
 },
\end{equation}
where $T > 0$ is a temperature parameter. This choice is motivated as follows~\cite{zhu00exploring}: since statistics are computed from spatial averages of filter responses, one can show that, in the limit of infinitely large lattices, the distribution $p(\x)$ is zero outside the Julesz ensemble and uniform inside. In this manner, \cref{e:pjul} can be though as a uniform distribution over images that have a certain characteristic filter responses $\bar\mu = (\bar \mu_1,\dots,\bar\mu_L)$. 

Note also that the texture is completely described by the filter bank $F=(F_1,\dots,F_L)$ and their characteristic responses $\bar\mu$. As discussed below, the filter bank is generally fixed, so in this framework different textures are given by different characteristics $\bar \mu$.

\paragraph{Generation-by-minimization.} For any interesting choice of the filter bank $F$, sampling from \cref{e:pjul} is rather challenging and classically addressed by Monte Carlo methods~\cite{zhu00exploring}. In order to make this framework more practical, Portilla and Simoncelli~\cite{portilla00a-parametric} proposed instead to heuristically sample from the Julesz ensemble by the optimization process
\begin{equation}\label{e:simo}
x^* = \operatornamewithlimits{argmin}_{x\in\mathcal{X}} \mathcal{L}(x).
\end{equation}
If this optimization problem can be solved, the minimizer $x^*$ is by definition a texture image. However, there is no reason why this process should generate fair samples from the distribution $p(x)$. In fact, the only reason why \cref{e:simo} may not simply return \emph{always the same} image is that the optimization algorithm is randomly initialized, the loss function is highly non-convex, and search is local. Only because of this \cref{e:simo} may land on different samples $x^*$ on different runs.

\paragraph{Deep filter banks.} Constructing a Julesz ensemble requires choosing a filter bank $F$. Originally, researchers considered the obvious candidates: Gaussian derivative filters, Gabor filters, wavelets, histograms, and similar~\cite{zhu00exploring,portilla00a-parametric,Zhu98}. More recently, the work of Gatys~\etal~\cite{Gatys15,Gatys16} demonstrated that much superior filters are automatically learned by deep convolutional neural networks (CNNs) even when trained for apparently unrelated problems, such as image classification. In this paper, in particular, we choose for $\mathcal{L}(x)$ the \emph{style loss} proposed by \cite{Gatys15}. The latter is the distance between the empirical correlation matrices of deep filter responses in a CNN.\footnote{Note that such matrices are obtained by averaging local non-linear filters: these are the outer products of filters in a certain layer of the neural network. Hence, the style loss of Gatys~\etal. is in the same form as~\cref{e:loss}.}

\paragraph{Stylization.} The texture generation method of Gatys \etal~\cite{Gatys15} can be considered as a direct extension of the texture generation-by-minimization technique~\eqref{e:simo} of Portilla and Simoncelli~\cite{portilla00a-parametric}. Later, Gatys \etal~\cite{Gatys16} demonstrated that the same technique can be used to generate an image that mixes the statistics of two other images, one used as a texture template and one used as a content template. Content is captured by introducing a second loss $\mathcal{L}_\text{cont.}(x,x_0)$ that compares the responses of deep CNN filters extracted from the generated image $x$ and a content image $x_0$. Minimizing the combined loss $\mathcal{L}(x)+\alpha\mathcal{L}_\text{cont.}(x,x_0)$ yields impressive \emph{artistic images}, where a texture $\bar \mu$, defining the artistic style, is fused with the content image $x_0$.
 
\paragraph{Feed-forward generator networks.} For all its simplicity and efficiency compared to Markov sampling techniques, generation-by-optimization~\eqref{e:simo} is still relatively slow, and certainly too slow for real-time applications. Therefore, in the past few months several authors~\cite{Johnson16,Ulyanov16} have proposed to \emph{learn generator neural networks} $g(\z)$ that can directly map random noise samples $\z \sim p_z = \mathcal{N}(0,I)$ to a local minimizer of \cref{e:simo}. Learning the neural network $g$ amounts to minimizing the objective
\begin{equation}\label{e:obj-gen}
g^* = \argmin_g \E_{p_z}\mathcal{L}(g(\z)).
\end{equation}
While this approach works well in practice, it shares the same important limitation as the original work of Portilla and Simoncelli: there is no guarantee that samples generated by $g^*$ would be fair samples of the texture distribution~\eqref{e:pjul}. In practice, as we show in the paper, such samples tend in fact to be not diverse enough.

Both~\cite{Johnson16,Ulyanov16} have also shown that similar generator networks work also for stylization. In this case, the generator $g(\x_0,\z)$ is a function of the content image $\x_0$ and of the random noise $\z$. The network $g$ is learned to minimize the sum of texture loss and the content loss:
\begin{equation}\label{e:obj-synth}
g^* = \argmin_g \E_{p_{\x_0},p_{\z}}[\mathcal{L}(g(\x_0,\z)) + \alpha \mathcal{L}_\text{cont.}(g(\x_0,\z),\x_0)].
\end{equation}

\paragraph{Alternative neural generator methods.} There are many other techniques for image generation using deep neural networks.

The Julesz distribution is closely related to the FRAME maximum entropy model of~\cite{Zhu98}, as well as to the concept of \emph{Maximum Mean Discrepancy} (MMD) introduced in~\cite{Gretton06}. Both FRAME and MMD make the observation that a probability distribution $p(\x)$ can be described by the expected values $\mu_\alpha = \mathbb{E}_{\x \sim p(\x)}[\phi_\alpha(\x)]$ of a sufficiently rich set of statistics $\phi_\alpha(x)$. Building on these ideas, \cite{Li15,Dziugaite15} construct generator neural networks $g$ with the goal of minimizing the discrepancy between the statistics averaged over a batch of generated images $\sum_{i=1}^N \phi_\alpha(g(\z_i))/N$ and the statistics averaged over a training set $\sum_{i=1}^M \phi_\alpha(x_i)/M$. The resulting networks $g$ are called  \emph{Moment Matching Networks} (MMN).

An important alternative methodology is based on the concept of Generative Adversarial Networks (GAN;~\cite{Goodfellow14}). This approach trains, together with the generator network $g(\z)$, a second \emph{adversarial} network $f(\cdot)$ that attempts to distinguish between generated samples $g(\z),\z\sim\mathcal{N}(0,I)$ and real samples $\x\sim p_\text{data}(\x)$. The adversarial model $f$ can be used as a measure of quality of the generated samples and used to learn a better generator $g$. GAN are powerful but notoriously difficult to train. A lot of research is has recently focused on improving GAN or extending it. For instance, LAPGAN~\cite{Denton15} combines GAN with a Laplacian pyramid and DCGAN~\cite{Radford15} optimizes GAN for large datasets.
\section{Julesz generator networks}\label{s:julesz}

This section describes our first contribution, namely a method to learn networks that draw samples from the Julesz ensemble modeling a texture (\cref{s:background}), which is an intractable problem usually addressed by slow Monte Carlo methods~\cite{Zhu98,zhu00exploring}. Generation-by-optimization, popularized by Portilla and Simoncelli and Gatys~\etal, is faster, but can only find one point in the ensemble, not sample from it, with scarce sample diversity, particularly when used to train feed-forward generator networks~\cite{Johnson16,Ulyanov16}.

Here, we propose a new formulation that allows to train generator networks that \emph{sample the Julesz ensemble}, generating images with high visual fidelity as well as high diversity. 

A generator network~\cite{Goodfellow14} maps an i.i.d. noise vector $\z\sim\mathcal{N}(0,I)$ to an image $x = g(\z)$ in such a way that $x$ is ideally a sample from the desired distribution $p(x)$. Such generators have been adopted for texture synthesis in~\cite{Ulyanov16}, but without guarantees that the learned generator $g(\z)$ would indeed sample a particular distribution.

Here, we would like to sample from the Gibbs distribution~\eqref{e:pjul} defined over the Julesz ensemble. This distribution can be written compactly as $p(x) = Z^{-1} e^{-\mathcal{L}(x)/T}$, where $Z = \int e^{-\mathcal{L}(x)/T}\,dx$ is an intractable normalization constant.

Denote by $q(x)$ the distribution induced by a generator network $g$. The goal is to make the target distribution $p$ and the generator distribution $q$ as close as possible by minimizing their Kullback-Leibler (KL) divergence:
\begin{align}\label{e:kl}
\begin{split}
KL(q||p) &= 
\int q(x)\ln \frac{q(x)Z}{p(x)}\,dx
\\
&=
\frac{1}{T} \E_{\x \sim q(\x)} \mathcal{L}(\x) + \E_{x \sim q(\x)} \ln q(\x) +  \ln(Z)
\\
&=
\frac{1}{T} \E_{\x \sim q(\x)} \mathcal{L}(\x) - H(q) + \text{const.}
\end{split}
\end{align}
Hence, the KL divergence is the sum of the expected value of the style loss $\mathcal{L}$ and the negative entropy of the generated distribution $q$.

The first term can be estimated by taking the expectation over generated samples:
\begin{equation}\label{e:exploss}
\E_{\x \sim q(\x)} \mathcal{L}(\x) =
\E_{\z \sim \mathcal{N}(0,I)} \mathcal{L}(g(\z)).
\end{equation}
This is similar to the reparametrization trick of~\cite{Kingma13} and is also used in~\cite{Johnson16,Ulyanov16} to construct their learning objectives.

The second term, the negative entropy, is harder to estimate accurately, but simple estimators exist. One which is particularly appealing in our scenario is the Kozachenko-Leonenko estimator~\cite{Kozachenko87}. This estimator considers a \emph{batch} of $N$ samples $x_1,\dots,x_n \sim q(x)$. Then, for each sample $x_i$, it computes the distance $\rho_i$ to its nearest neighbour in the batch:
\begin{equation}
\rho_i = \min_{j\ne i} \| x_i - x_j \|.
\end{equation}
The distances $\rho_i$ can be used to approximate the entropy as follows:
\begin{equation}\label{e:entest}
H(q) \approx \frac{D}{N} \sum_{i=1}^N \ln \rho_i + \text{const.}
\end{equation}
where $D=3WH$ is the number of components of the images $x \in \mathbb{R}^{3 \times W \times H}$.

An energy term similar to~\eqref{e:kl} was recently proposed in~\cite{kim16deep} for improving the diversity of a generator network in a adversarial learning scheme. While the idea is superficially similar, the application (sampling the Julesz ensemble) and instantiation (the way the entropy term is implemented) are very different.

\paragraph{Learning objective.} We are now ready to define an objective function $E(g)$ to learn the generator network $g$. This is given by substituting the expected loss~\eqref{e:exploss} and the entropy estimator~\eqref{e:entest}, computed over a batch of $N$ generated images, in the KL divergence~\eqref{e:kl}:
\begin{multline}\label{e:jloss}
E(g) =
  \frac{1}{N}
  \sum_{i=1}^N
  \Big[
  \frac{1}{T}\mathcal{L}(g(\z_i))
  \\
  -
  \lambda
  \ln \min_{j\ne i}
  \| g(\z_i) - g(\z_j) \|
  \Big]
\end{multline}
The batch itself is obtained by drawing $N$ samples $\z_1,\dots,\z_n \sim \mathcal{N}(0,I)$ from the noise distribution of the generator. The first term in~\cref{e:jloss} measures how closely the generated images $g(\z_i)$ are to the Julesz ensemble. The second term quantifies the lack of diversity in the batch by mutually comparing the generated images.

\paragraph{Learning.} The loss function~\eqref{e:jloss} is in a form that allows optimization by means of Stochastic Gradient Descent (SGD). The algorithm samples a batch $\z_1,\dots,\z_n$ at a time and then descends the gradient:
\begin{multline}
\frac{1}{N}\sum_{i=1}^N
  \Big[
  \frac{d\mathcal{L}}{dx^\top} \frac{d g(\z_i)}{d\theta^\top}
  \\
  -
  \frac{\lambda}{\rho_i}
  (g(\z_i) - g(\z_{j^*_i}))^\top
  \left(
  \frac{dg(\z_i)}{d\theta^\top}
  -
  \frac{dg(\z_{j^*_i})}{d\theta^\top}
  \right)
  \Big]
\end{multline}
where $\theta$ is the vector of parameters of the neural network $g$, the tensor image $x$ has been implicitly vectorized and $j^*_i$ is the index of the nearest neighbour of image $i$ in the batch.

\section{Stylization with instance normalization}\label{s:instance}

The work of~\cite{Ulyanov16} showed that it is possible to learn high-quality texture networks $g(\z)$ that generate images in a Julesz ensemble. They also showed that it is possible to learn good quality stylization networks $g(x_0,z)$ that apply the style of a fixed texture to an arbitrary content image $x_0$. 

Nevertheless, the stylization problem was found to be harder than the texture generation one. For the stylization task, they found that learning the model from too many example content images $x_0$, say more than 16, yielded poorer \emph{qualitative} results than using a smaller number of such examples. Some of the most significant errors appeared along the border of the generated images, probably due to padding and other boundary effects in the generator network. We conjectured that these are symptoms of a learning problem too difficult for their choice of neural network architecture.

A simple observation that may make learning simpler is that the result of stylization should not, in general, depend on the contrast of the content image 
but rather should match the contrast of the texture that is being applied to it. Thus, the generator network should discard contrast information in the content image $x_0$. We argue that learning to discard contrast information by using standard CNN building block is unnecessarily difficult, and is best done by adding a suitable layer to the architecture.

To see why, let $x\in\mathbb{R}^{N \times C \times W \times H}$ be an input tensor containing a batch of $N$ images. Let $x_{nijk}$ denote its $nijk$-th element, where $k$ and $j$ span spatial dimensions, $i$ is the feature channel (i.e.\ the color channel if the tensor is an RGB image), and $n$ is the index of the image in the batch. Then, contrast normalization is given by: 
\begin{align}\label{eq:inorm}
\begin{split}
y_{nijk} &= \frac{x_{nijk} - \mu_{ni}}{\sqrt{\sigma_{ni}^2 + \epsilon}},
\\
\mu_{ni} &= \frac{1}{HW}\sum_{l=1}^W \sum_{m=1}^H x_{nilm},
\\
\sigma_{ni}^2 &= \frac{1}{HW}\sum_{l=1}^W \sum_{m=1}^H (x_{nilm} - \mu_{ni})^2.
\end{split}
\end{align}
It is unclear how such as function could be implemented as a sequence of standard operators such as ReLU and convolution.

On the other hand, the generator network of~\cite{Ulyanov16} does contain a normalization layers, and precisely \emph{batch normalization} (BN) ones. The key difference between \cref{eq:inorm} and batch normalization is that the latter applies the normalization to a whole batch of images instead of single ones:
\begin{align}\label{eq:bnorm}
\begin{split}
    y_{nijk} &=  \frac{x_{nijk} - \mu_{i}}{\sqrt{\sigma_i^2 + \epsilon}},
    \\
    \mu_i &= \frac{1}{HWN}\sum_{n=1}^N\sum_{l=1}^W \sum_{m=1}^H x_{nilm},
    \\
    \sigma_i^2 &= \frac{1}{HWN}\sum_{n=1}^N\sum_{l=1}^W \sum_{m=1}^H (x_{nilm} - \mu_i)^2.
\end{split}
\end{align}
We argue that, for the purpose of stylization, the normalization operator of~\cref{eq:inorm} is preferable as it can normalize each individual content image $x_0$.

\begin{figure}
\centering
\includegraphics[width=1\linewidth]{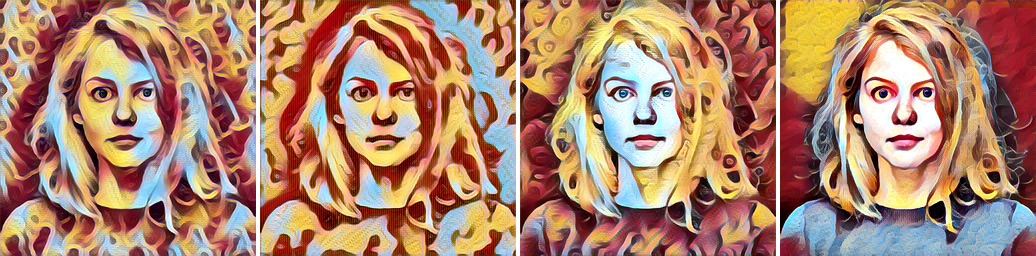}
\vspace{-2em}
\caption{Comparison of normalization techniques in image stylization. From left to right: BN, cross-channel LRN at the first layer, IN at the first layer, IN throughout.}
\label{fig:style_examples}
\end{figure}

While some authors call layer~\cref{eq:inorm} contrast normalization, here we refer to it as \emph{instance normalization} (IN) since we use it as a drop-in replacement for batch normalization operating on individual instances instead of the batch as a whole. Note in particular that this means that instance normalization is applied throughout the architecture, not just at the input image---\cref{fig:style_examples} shows the benefit of doing so.

\begin{figure}[t]
    \centering
    \hspace*{-2mm}\begin{subfigure}[b]{0.5\linewidth}
        \includegraphics[width=\linewidth]{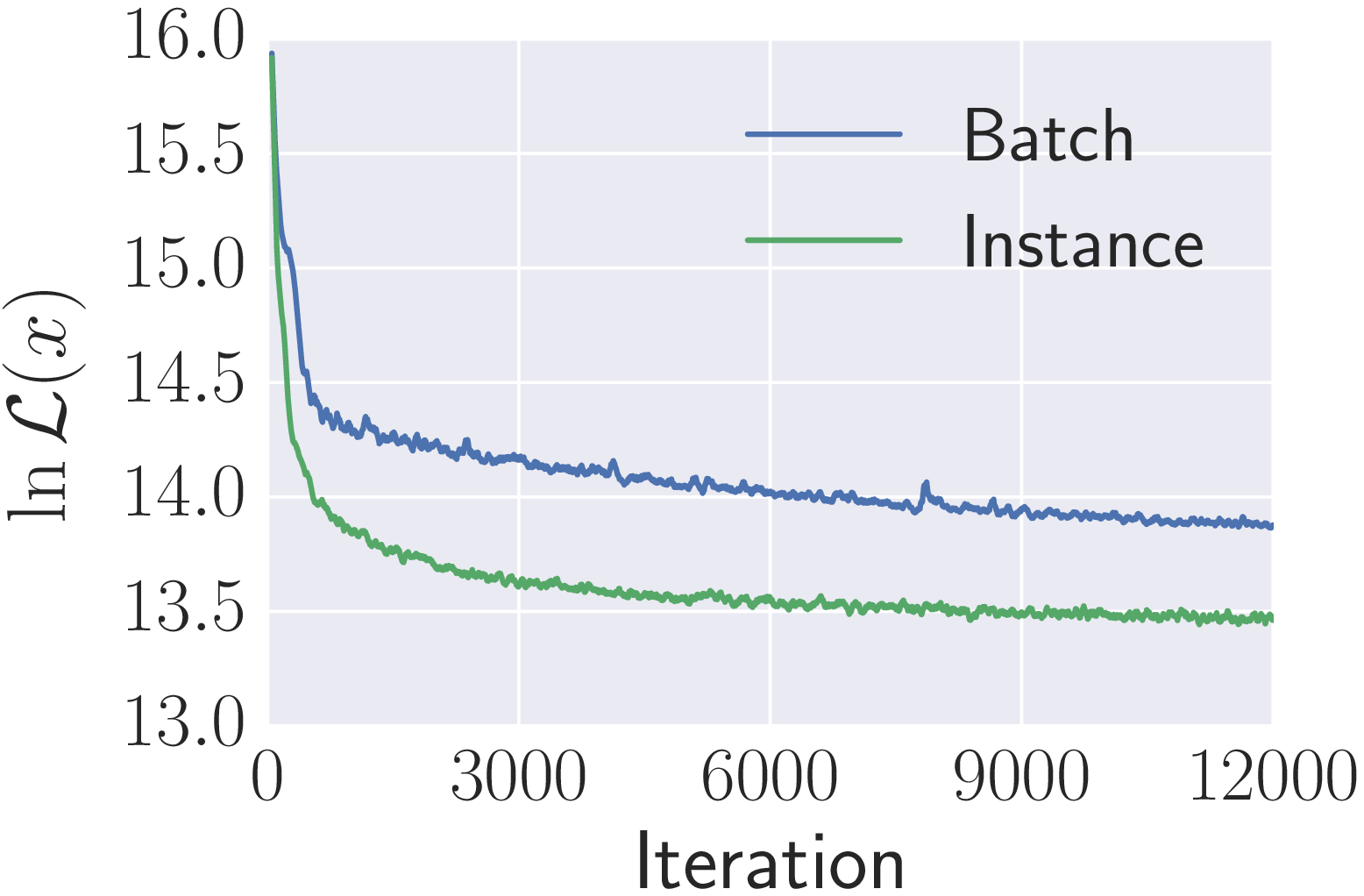}
        \vspace*{-6mm}\caption{Feed-forward history.}
    \end{subfigure}
    \begin{subfigure}[b]{0.49\linewidth}
        \includegraphics[width=\linewidth]{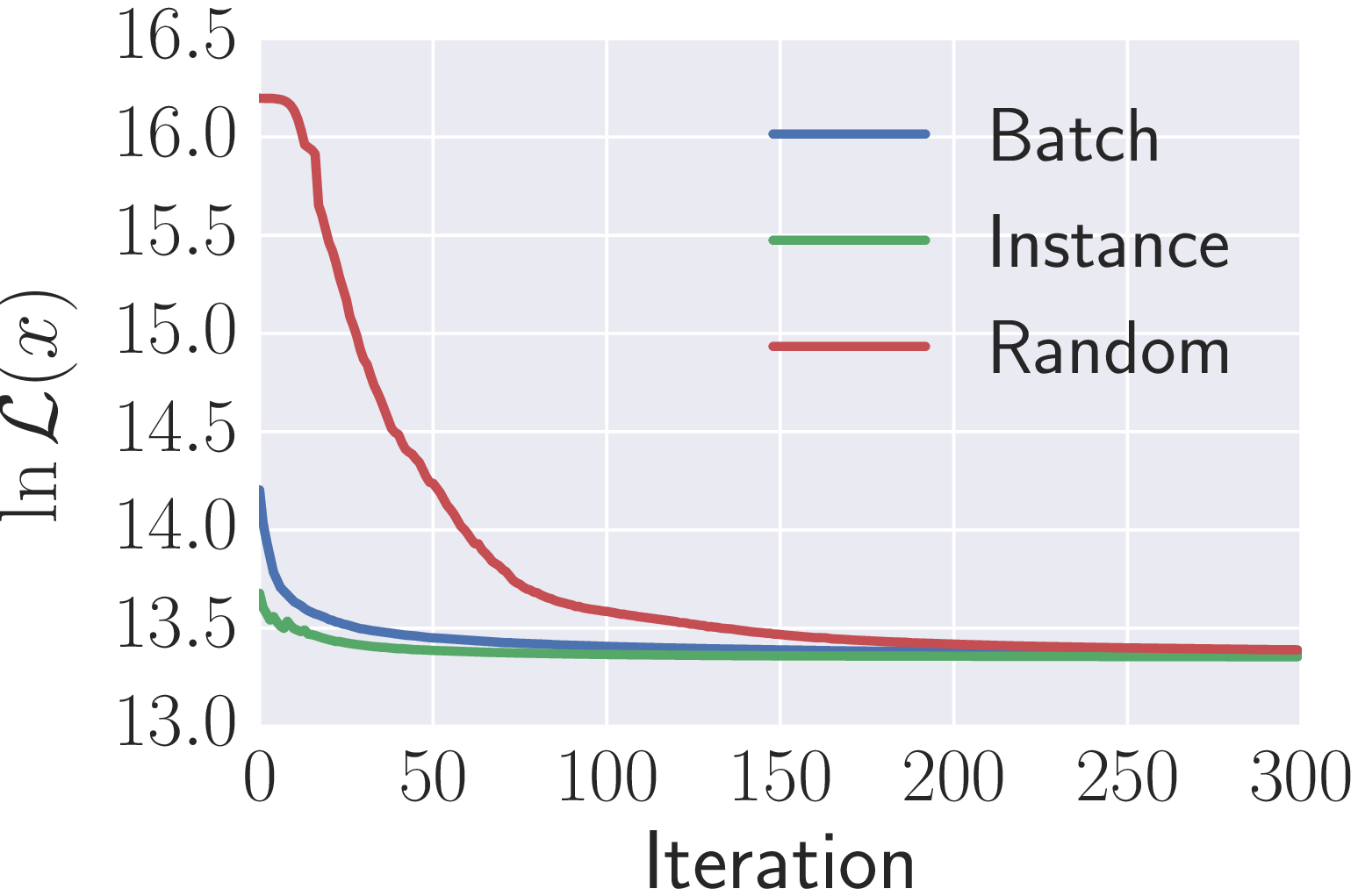}
        \vspace*{-6mm}\caption{Finetuning history.}
    \end{subfigure}
    \\
    \begin{subfigure}[b]{0.32\linewidth}
        \includegraphics[width=\linewidth]{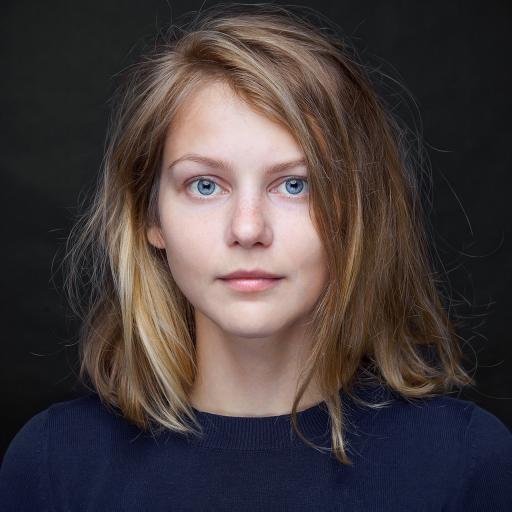}
        \vspace*{-6mm}\caption{Content.}
    \end{subfigure}
    \begin{subfigure}[b]{0.32\linewidth}
        \includegraphics[width=\linewidth]{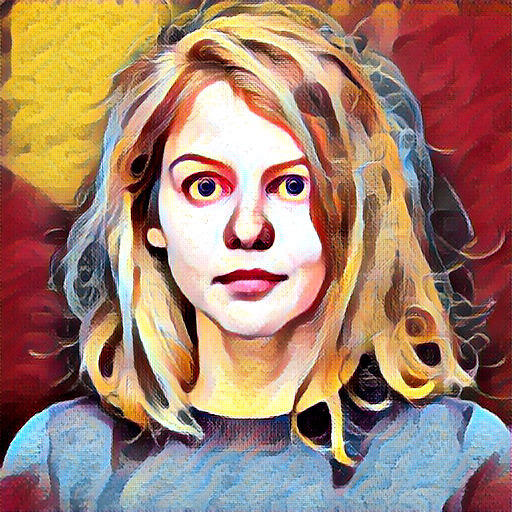}
        \vspace*{-6mm}\caption{StyleNet IN.}
    \end{subfigure}
    \begin{subfigure}[b]{0.32\linewidth}
        \includegraphics[width=\linewidth]{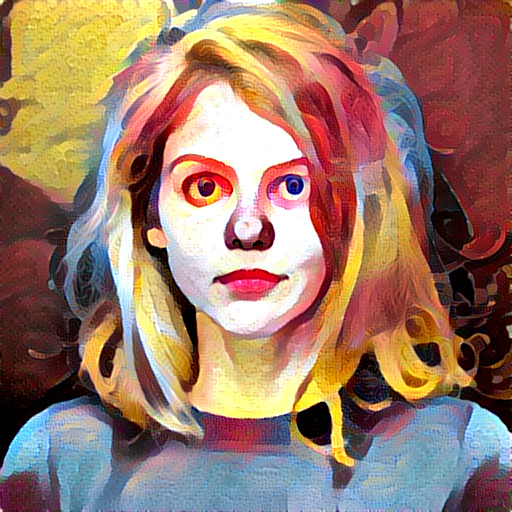}
        \vspace*{-6mm}\caption{IN finetuned.}
    \end{subfigure}
    \\
    \begin{subfigure}[b]{0.32\linewidth}
        \includegraphics[width=\linewidth]{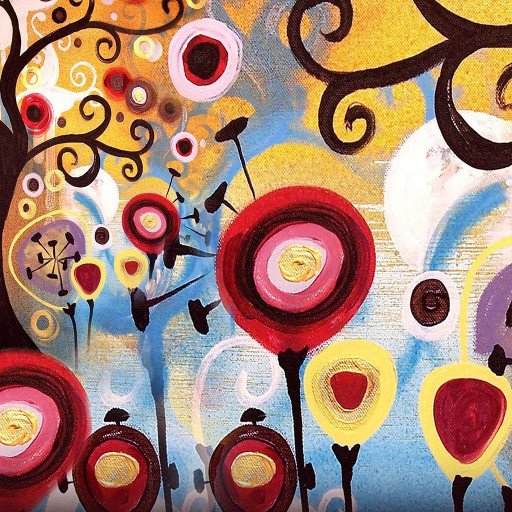}
        \vspace*{-6mm}\caption{Style.}
    \end{subfigure}
    \begin{subfigure}[b]{0.32\linewidth}
        \includegraphics[width=\linewidth]{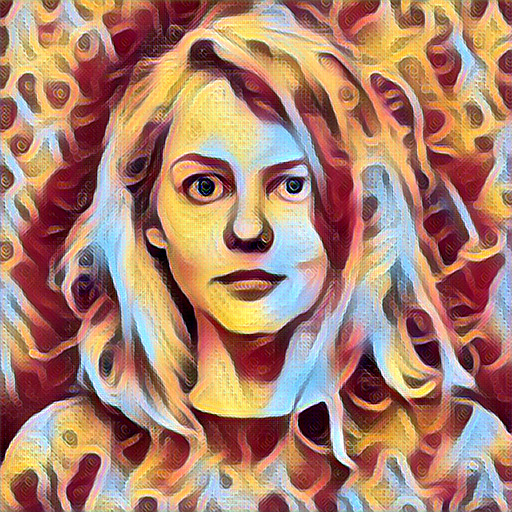}
        \vspace*{-6mm}\caption{StyleNet BN.}
    \end{subfigure}
    \begin{subfigure}[b]{0.32\linewidth}
        \includegraphics[width=\linewidth]{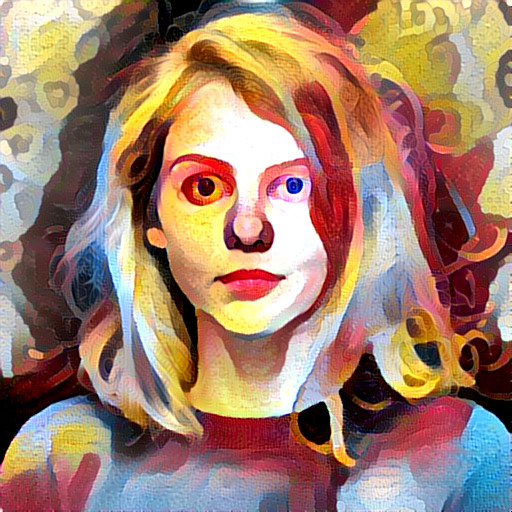}
        \vspace*{-6mm}\caption{BN finetuned.}
    \end{subfigure}
    \caption{(a) learning objective as a function of SGD iterations for StyleNet IN and BN. (b) Direct optimization of the Gatys~\etal for this example image starting from the result of StyleNet IN and BN. (d,g) Result of StyleNet with instance (d) and batch normalization (g). (e,h) Result of finetuing the Gatys~\etal energy.}\label{f:finetuning}
\end{figure}

Another similarity with BN is that each IN layer is followed by a scaling and bias operator $s \odot \bx + b$. A difference is that the IN layer is applied at test time as well, unchanged, whereas BN is usually switched to use accumulated mean and variance instead of computing them over the batch.

IN appears to be similar to the layer normalization method introduced in~\cite{BaKH16} for recurrent networks, although it is not clear how they handle spatial data. Like theirs, IN is a generic layer, so we tested it in classification problems as well. In such cases, it still work surprisingly well, but not as well as batch normalization (e.g.\ AlexNet~\cite{Krizhevsky12} IN has 2-3\% worse top-1 accuracy on ILSVRC~\cite{ILSVRC15} than AlexNet BN).
\begin{figure*}[t]
\deflen{fivelen}{0.16\textwidth}
\newcolumntype{Y}{>{\centering\arraybackslash}X}
\centering\resizebox{.9\textwidth}{!}{
\begin{minipage}{\textwidth}
\includegraphics[width=\textwidth]{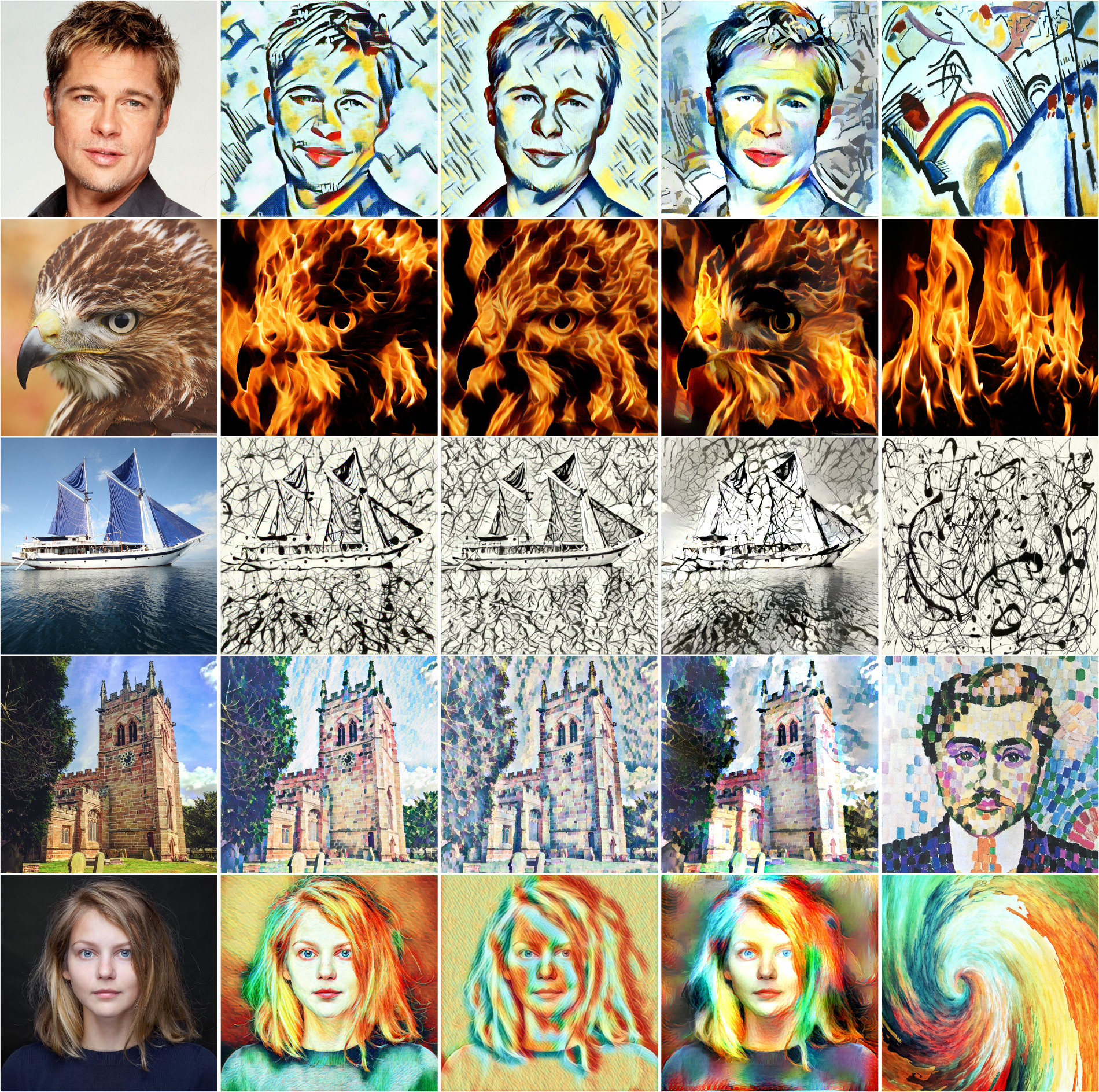}
\begin{tabularx}{\textwidth}{YYYYYY}
Content & 
\underline{StyleNet IN (ours)} & 
StyleNet BN & 
Gatys \etal & 
Style
\end{tabularx}
\end{minipage}
}
\vspace{-0.5em}
\caption{Stylization results obtained by applying different textures (rightmost column) to different content images (leftmost column). Three methods are compared: StyleNet IN, StyleNet BN, and iterative optimization.}
\label{fig:short}
\end{figure*}

\begin{figure*}[t]
\centering\resizebox{.9\textwidth}{!}{
\begin{minipage}{\textwidth}
\deflen{sixlen}{0.20\textwidth}
\deflen{firstlen}{0.02\textwidth}
\newcolumntype{A}{>{\centering\arraybackslash}p{0.1428\textwidth}}
\newcolumntype{B}{>{\centering\arraybackslash}p{0.2856\textwidth}}
\includegraphics[width=1.0\textwidth]{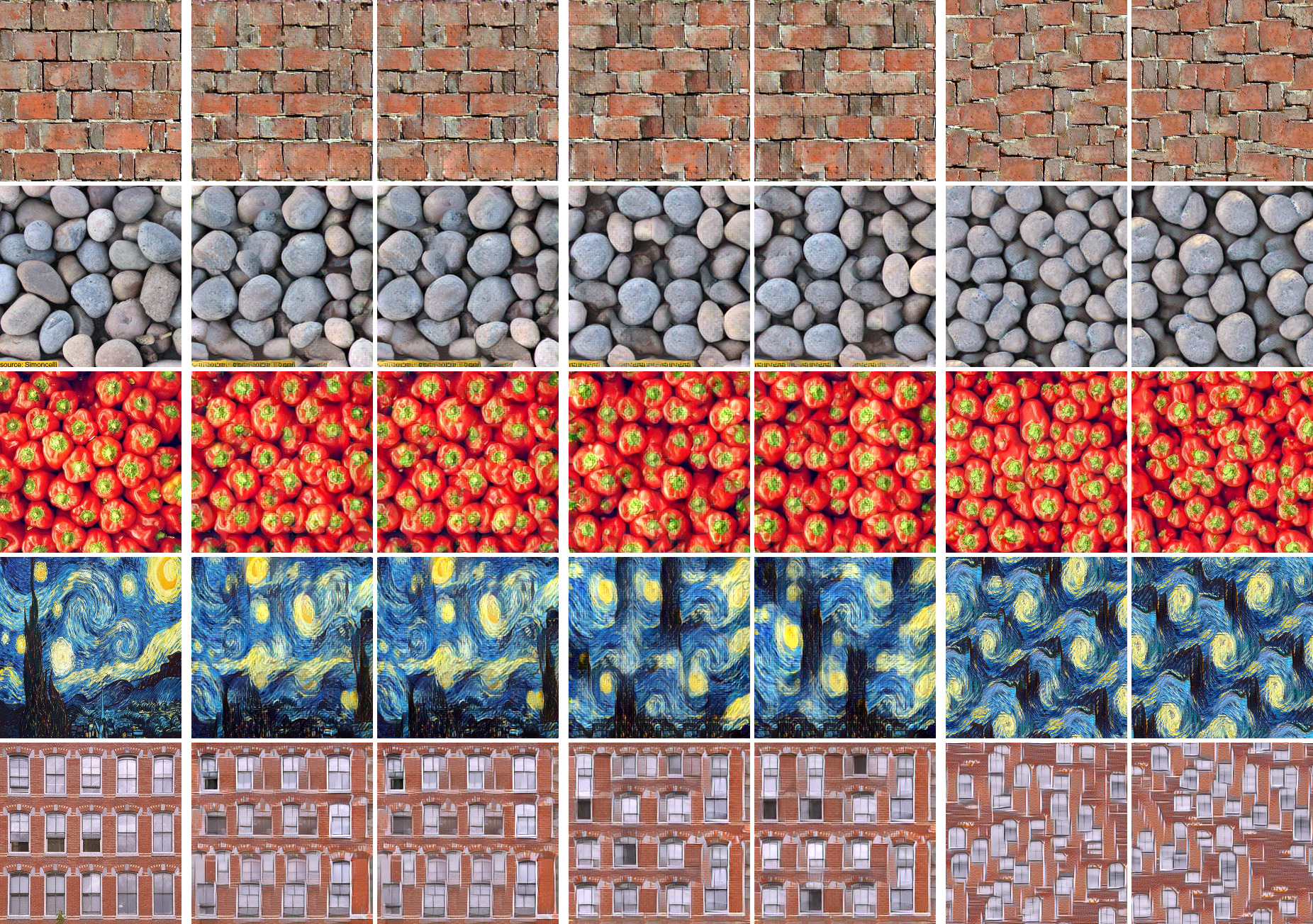}\\
\setlength{\tabcolsep}{0pt}
    \begin{tabularx}{\textwidth}{ABBB}
    Input &
    TextureNetV2 ~~$\lambda=0$ &
    \underline{TextureNetV2 ~~$\lambda>0$ (ours)} &
    TextureNetV1 ~~$\lambda=0$
    \end{tabularx}
\end{minipage}
}
\vspace{-0.5em}
\caption{The textures generated by the high capacity Texture Net V2 without diversity term ($\lambda =0$ in \cref{e:jloss}) are nearly identical. The low capacity TextureNet V1 of~ \cite{Ulyanov16} achieves diversity, but has sometimes poor results. TextureNet V2 with diversity is the best of both worlds.}
\label{f:texture_diversity}
\end{figure*}

\section{Experiments}\label{s:experiments}

In this section, after discussing the technical details of the method, we  evaluate our new texture network architectures using instance normalization, and then investigate the ability of the new formulation to learn diverse generators.

\subsection{Technical details}\label{s:technical}

\paragraph{Network architecture.} Among two generator network architectures, proposed previously in ~\cite{Ulyanov16,Johnson16}, we choose the residual architecture from~\cite{Johnson16} for all our style transfer experiments. We also experimented with architecture from~\cite{Ulyanov16} and observed a similar improvement with our method, but use the one from~\cite{Johnson16} for convenience. We call it \textit{StyleNet} with a postfix BN if it is equipped with batch normalization or IN for instance normalization.

For texture synthesis we compare two architectures: the multi-scale fully-convolutional architecture from~\cite{Ulyanov16} (TextureNetV1) and the one we design to have a very large receptive field (TextureNetV2). TextureNetV2 takes a noise vector of size $256$ and first transforms it with two fully-connected layers. The output is then reshaped to a $4 \times 4$ image and repeatedly upsampled with fractionally-strided convolutions similar to~\cite{Radford15}. More details can be found in the supplementary material. 

\begin{figure*}[t]
\centering\resizebox{.9\textwidth}{!}{
\begin{minipage}{\textwidth}
\newcolumntype{A}{>{\centering\arraybackslash}p{0.209\textwidth}}
\newcolumntype{D}{>{\centering\arraybackslash}p{0.1656\textwidth}}
\newcolumntype{B}{>{\centering\arraybackslash}p{0.220\textwidth}}
\newcolumntype{C}{>{\centering\arraybackslash}p{0.420\textwidth}}
\includegraphics[width=\linewidth]{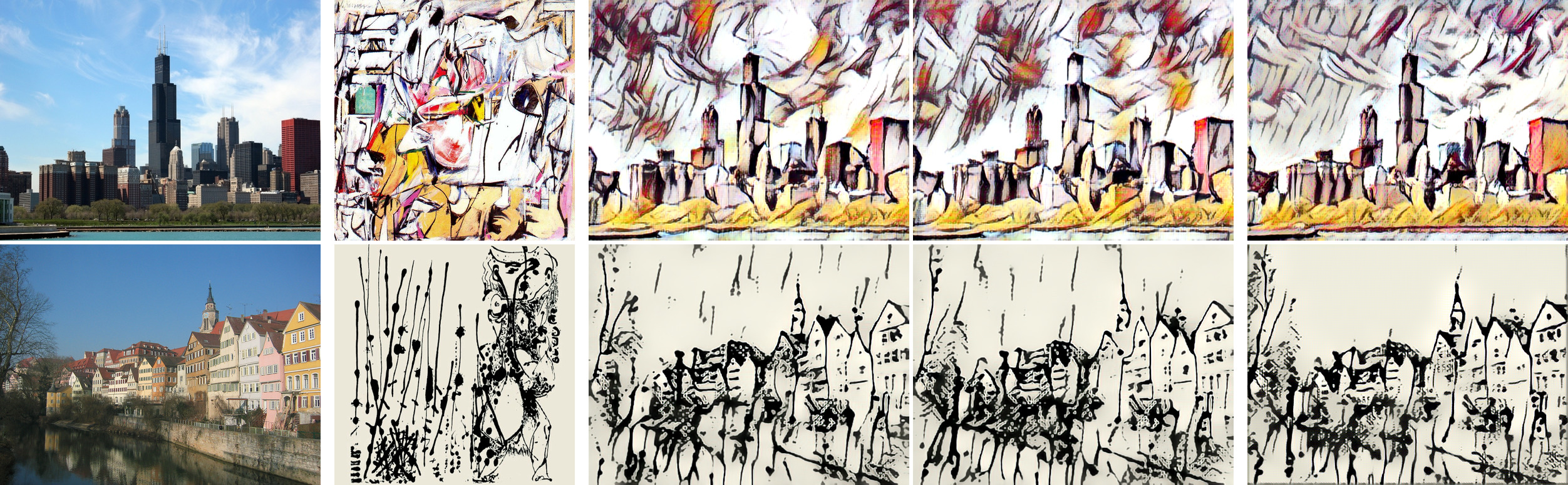}\\
\setlength{\tabcolsep}{0pt}
    \begin{tabularx}{\textwidth}{ADCB}
    Content &
    Style &
    StyleNet ~~$\lambda>0$&
    StyleNet $\lambda=0$
    \end{tabularx}
\end{minipage}
}
\vspace{-0.5em}
\caption{The StyleNetV2 $g(x_0,z)$, trained with diversity $\lambda > 0$, generates substantially different stylizations for different values of the input noise $z$. In this case, the lack of stylization diversity is visible in uniform regions such as the sky.}\label{fig:stylevar}
\vspace{0.5em}
\end{figure*}

\begin{figure}[t]
\centering\includegraphics[width=0.7\linewidth]{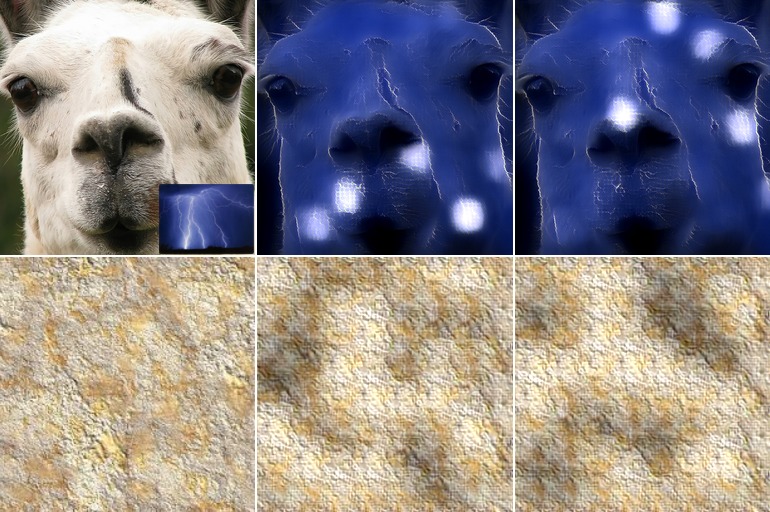}
\caption{Negative examples. If the diversity term $\lambda$ is too high for the learned style, the generator tends to generate artifacts in which brightness is changed locally (spotting) instead of (or as well as) changing the structure.}\label{f:negative}
\end{figure}

\paragraph{Weight parameters.}  In practice, for the case of $\lambda>0$, entropy loss and texture loss in \cref{e:jloss} should be weighted properly. As only the value of $T\lambda$ is important for optimization we assume $\lambda=1$ and choose $T$ from the set of three values $(5,10,20)$ for texture synthesis (we pick the higher value among those not leading to artifacts -- see our discussion below). We fix $T=10000$ for style transfer experiments. For texture synthesis, similarly to~\cite{Ulyanov16}, we found useful to normalize gradient of the texture loss as it passes back through the VGG-19 network. This allows rapid convergence for stochastic optimization but implicitly alters the objective function and requires temperature to be adjusted. We observe that for textures with flat lightning high entropy weight results in brightness variations over the image \cref{f:negative}. We hypothesize this issue can be solved if either more clever distance for entropy estimation is used or an image prior introduced.

\subsection{Effect of instance normalization}\label{s:exp-instance}

In order to evaluate the impact of replacing batch normalization with instance normalization, we consider first the problem of \emph{stylization}, where the goal is to learn a generator $x=g(x_0,z)$ that applies a certain texture style to the content image $x_0$ using noise $z$ as ``random seed''. We set $\lambda=0$ for which generator is most likely to discard the noise. 

The StyleNet IN and StyleNet BN are compared in \cref{f:finetuning}. Panel \cref{f:finetuning}.a shows the training objective \eqref{e:obj-synth} of the networks as a function of the SGD training iteration. The objective function is the same, but StyleNet IN converges much faster, suggesting that it can solve the stylization problem more easily. This is confirmed by the stark difference in the qualitative results in panels (d) end (g). Since the StyleNets are trained to minimize in one shot the same objective as the iterative optimization of Gatys~\etal., they can be used to \emph{initialize} the latter algorithm. Panel (b) shows the result of applying the Gatys~\etal optimization starting from their random initialization and the output of the two StyleNets. Clearly both networks start much closer to an optimum than random noise, and IN closer than BN. The difference is qualitatively large: panels (e) and (h) show the change in the StyleNets output after finetuning by iterative optimization of the loss, which has a small effect for the IN variant, and a much larger one for the BN one.

Similar results apply in general. Other examples are shown in \cref{fig:short}, where the IN variant is far superior to BN and much closer to the results obtained by the much slower iterative method of Gatys~\etal. StyleNets are trained on images of a fixed sized, but since they are convolutional, they can be applied to arbitrary sizes. In the figure, the top tree images are processed at $512\times 512$ resolution and the bottom two at $1024\times 1024$. In general, we found that higher resolution images yield visually better stylization results. 

While instance normalization works much better than batch normalization for stylization, for texture synthesis the two normalization methods perform equally well. This is consistent with our intuition that IN helps in normalizing the information coming from content image $x_0$, which is highly variable, whereas it is not important to normalize the texture information, as each model learns only one texture style.

\subsection{Effect of the diversity term}\label{s:exp-diversity}

Having validated the IN-based architecture, we evaluate now the effect of the entropy-based diversity term in the objective function~(\ref{e:jloss}). 

The experiment in~\cref{f:texture_diversity} starts by considering the problem of texture generation. We compare the new high-capacity TextureNetV2 and the low-capacity TextureNetsV1 texture synthesis networks. The low-capacity model is the same as~\cite{Ulyanov16}. This network was used there in order to force the network to learn a non-trivial dependency on the input noise, thus generating diverse outputs even though the learning objective of~\cite{Ulyanov16}, which is the same as~\cref{e:jloss} with diversity coefficient $\lambda=0$, tends to suppress diversity. The results in~\cref{f:texture_diversity} are indeed diverse, but sometimes of low quality. This should be contrasted with TextureNetV2, the high-capacity model: its visual fidelity is much higher, but, by using the same objective function~\cite{Ulyanov16}, the network learns to generate a single image, as expected. TextureNetV2 with the new diversity-inducing objective ($\lambda > 0$) is the best of both worlds, being both high-quality and diverse.

The experiment in~\cref{fig:stylevar} assesses the effect of the diversity term in the stylization problem. The results are similar to the ones for texture synthesis and the diversity term effectively encourages the network to learn to produce different results based on the input noise.

One difficultly with texture and stylization networks is that the entropy loss weight $\lambda$ must be tuned for each learned texture model. Choosing $\lambda$ too small may fail to learn a diverse generator, and setting it too high may create artifacts, as shown in~\cref{f:negative}.

\section{Summary}\label{s:conclusions}

This paper advances feed-forward texture synthesis and stylization networks in two significant ways. It introduces instance normalization, an architectural change that makes training stylization networks easier and allows the training process to achieve much lower loss levels. It also introduces a new learning formulation for training generator networks to sample uniformly from the Julesz ensemble, thus explicitly encouraging diversity in the generated outputs. We show that both improvements lead to noticeable improvements of the generated stylized images and textures, while keeping the generation run-times intact.

{\small
\bibliographystyle{ieee}
\bibliography{egbib}}
\end{document}